# Predicting Network Attacks Using Ontology-Driven Inference

Ahmad Salahi

Information and Communication Security Department
Research Institute for ICT (ex ITRC)
Tehran, Iran
salahi@itrc.ac.ir

Morteza Ansarinia

Information and Communication Security Department
Research Institute for ICT (ex ITRC)
Tehran, Iran
ansarinia@me.com

*Abstract*— **Graph knowledge models and ontologies are very powerful modeling and reasoning tools. We propose an effective approach to model network attacks and attack prediction which plays important roles in security management. The goals of this study are: First we model network attacks, their prerequisites and consequences using knowledge representation methods in order to provide description logic reasoning and inference over attack domain concepts. And secondly, we propose an ontology-based system which predicts potential attacks using inference and observing information which provided by sensory inputs. We generate our ontology and evaluate corresponding methods using CAPEC, CWE, and CVE hierarchical datasets. Results from experiments show significant capability improvements comparing to traditional hierarchical and relational models. Proposed method also reduces false alarms and improves intrusion detection effectiveness.**

*Keywords- Knowledge Engineering, Network Security, Ontology.*

## I. Introduction

Knowledge representation is an important technique in the context of the security management, since it provides a mechanism via which heterogeneous systems can interact with each other using same semantic, but different syntax. It also provides machines with powerful semantic–level capabilities such as logical inference without any complex prerequisite.

In knowledge engineering, graph models are powerful data storages, which represent more concrete knowledge of a domain comparing to relational databases and hierarchical taxonomies. Ontology is a graph model represents knowledge of a domain by which developers and machines can share information in the domain. Ontologies mainly include concepts in the domain and relations among the concepts. It is a standard method to share common understanding of domain among experts, reuse domain knowledge, make assumptions explicit, separate domain knowledge from operational knowledge, and finally analyze domain of interest knowledge by means of logic.

Despite various classifications and categorizations of network attacks, there is still no agreement on their structure and formal nature. To make some progress towards a shared conceptualization of network attacks, and using this conceptualization as a method to predict potential attacks, this paper presents a general models of network attacks, and describes a method by which potential attacks can be predicted using inference over the network attacks model (which is a semantic representation of data received from input intrusion detection sensors). From representation perspective. we focus on how an ontology and corresponding knowledge base could be constructed from informal and semi-informal data sources; and from problem solving perspective of ontologies, we use formal definitions to predict possible attacks in a system or a network. Predicting a network attack prior to the actual incident helps to resolve vulnerabilities, update system configurations and fix weaknesses, prevent consequences like data, and last but not least prevent potential attacks to happen based on our knowledge of the current state.

Our contribution is to not only transform information about network attacks from being machine-readable to machine-understandable, but also use produced knowledge to predict potential attacks according to the vulnerabilities, weaknesses, and prerequisites of an attack. Moreover proposed ontology can be used as a common vocabulary and semantic representation of attacks for agents in the network (e.g. IDSs), by which machines can communicate.

While our concern is to design an ontology and more importantly to use the ontology as a base for further reasoning techniques, it's desired to have a consistent, extendable, and coherent ontology. Section VI explores some evaluation techniques to check these criteria. Figure 1 illustrates overall architecture of proposed system. For the purpose of standardizing logs and event, CEE dataset [25] (a unified event classification by combining support for multiple event syntaxes and log protocols) is used.

We first provide a brief review of researches about network attack definitions, followed by the definition of ontology as a model for semantics in this domain. Section IV describes a methodology by which a powerful and descriptive ontology on network attack can be constructed by using some semi-informal sources. In section V we propose a method to predict possible scenarios in which a network attack could happen. These scenarios are direct results of machine inferences over underlying ontology constructed in the previous step. Outputs of such description logic inferences could be weaknesses, vulnerabilities and potential attack patterns that have to be taken care of. We conclude in Section VII.

## II. BACKGROUND

In this section, we first provide a review of current researches of representing and predicting network attacks, including their limitations. And second part of this section briefly describes ontologies and our desired properties of an ontology for the task of network attack prediction.

## LITERATURE REVIEW

Formal definition of network attacks is mostly limited to taxonomies and attack languages. Ontological modeling of attacks is a new category of research which has few published research papers in the literature. Reference [29] introduces an ontology for knowledge representation goal, and its applications for information security are investigated. References [23] and [30] also provide modeling for intrusion detection, and a limited ontology of well-known attacks is introduced.

The majority of existing resources in the area of prediction and ontological modeling of network attacks are limited to biology and prediction in the domain of genes, biological entities. Below is an exploration on taxonomies, attack languages, and overall idea of prediction.

Various hierarchical data modeling and different taxonomies proposed for network attacks lead to a very large group of variant syntax representation of network attacks. OSBVDB [1] and Capec [5] both present taxonomies classified from the view of attacked entity. Moreover, reference [17] represents another taxonomy, in which attacks are classified based on how, when, and where they happen. We incorporated attacked entity's view into our ontology. In addition to attack categorizations, vulnerabilities and weaknesses are also categorized with regard to their relation to the attacks, respectively, in CVE [7] and CWE [20]. These two concepts (vulnerability and weakness) are also incorporated into our ontology for prediction purposes. In addition to these main concepts, [14][16][19] review consequences of attacks. According to [26], consequences are divided into *"(Unauthorized) Disclosure"*, *"Deception"*, *"Disruption"*, and *"Usurpation"* categories. These categories are also imported into our ontology for the reporting purposes of prediction, which is beyond focus of this paper. In another view of attacks, since IDSs are either adjacent to or co–located with attack targets, taxonomies are usually from attack target entities' view. We use this idea to model attack–related concepts, and predict attacks based on sensed logs and events. Additionally weaknesses, consequences, vulnerabilities, and other means of attacks are observed in [4], which are consistent with our ontology. On the contrary, [3] and [21] suggest modeling attack behavior from attacker's view in a taxonomical order. Reference [3] also states that attack detection is a young area of research, yet need more study to build an accepted framework. According to this statement, since ontologies work on semantic level rather than syntax level, developing an ontology for attack, inspired by the prediction task seems reasonable. As an outcome, ontologies can be used on growing taxonomy or attack language–based intrusion–related heterogeneous IDSs which use same concepts and yet different syntax as a common vocabulary.

Although taxonomies provide more information than simple list of concepts, they lack more interrelations between concepts and formal inferences. However, our proposed ontology is constructed upon existing attack taxonomy known as Capec, vulnerability taxonomy called CVE, and weakness taxonomy called CWE, all provided by Mitre. Our main contributions to the current state of these modeling and attack research domain are to discover inter–relationships and employ semantic–based methods to explore more intelligent techniques in the domain of discourse.

Taxonomies are not the only way to model network attacks. There are also attack languages in literature, classified as event reporting, correlation and recognition languages [10][11]. Compared to their possible features, ontologies are capable of simultaneously function similar to what reporting, correlating, and recognizing languages do.

Among various attack languages [18][13][22][24], STATL [10][11] is a finite–state machine based attack detection language, designed to describe an attack as a sequence of required actions or steps that need to be performed prior to the attack. Although STATL is an attacker–centric representation, we used such an attack pattern to describe multi–steps attacks in the attack prediction module (from IDSs' perspective) using semantic rules (using SWRL). STATL has some downsides like its lack of features to define larger attacks by merging minor attacks, vulnerabilities, and weaknesses.

CIDF [6] was also introduces in an effort to define protocols and interfaces to share attack information between researchers. The standard format of generalized intrusion detection object, which is core object to CIDF to exchange attack–related data, is defined in CIDL (Common Intrusion Detection Language) [12] which is capable of defining reports on the events.

The most active exchange format in the domain of network attack is IDMEF [8][31]. IDMEF uses XML to exchange data. IDMEF is an effort to establish a data model which defines computer intrusions. It defines a data model that represents data produced by an IDS. This data format is used in our system architecture to receive logs and events generated by IDSs, and then convert those sensed data into triples that are ready to be asserted into the knowledge base.

Modeling attack aside, detecting or predicting attack patterns are both desired goals for security managers. To predict or detect attack plans various statistical, probabilistic [33], analytical [34], and machine learning techniques like SVM [37][39], neural networks [39], and Genetic Algorithm [37] have been introduces [34]. In [36] active support vector domain description algorithm could find over 95% of predefined attacks. In general, machine learning methods lack invariant modeling for different representations of a same attack, and convergence problem (mostly in weak search algorithms like evolutionary techniques). They are also subjected to lack of extendibility for new attacks, vulnerabilities, and weaknesses. In addition, those methods suffer from too much parameters and process-consuming phase of optimization prior to be applicable to attack prediction or detection.

Comparing to the current recognition techniques, formal representation of attacks not only helps us to use common and straightforward logic and formal models instead of variant representations, but also by increasing efficiency which caused by too much parameters (system parameters which are used for anomaly analysis), provides faster response to security problems. Unlike machine learning algorithms, logical reasoning algorithms like forward-chaining and backward chaining seem sufficient. This leads to simple yet powerful solution. By introducing new concepts, their relations, and logical rules, formal knowledge makes it possible to resolve slow improvements in parameter optimization, and time-consuming training phase. Additionally, due to common representations of attacks for various tasks (e.g. prediction, detection, and reporting), ontological representation of computer attacks can be extended to new applications. References [35] and [38] present more in-depth reviews of various intrusion and anomaly detection techniques.

THEORY

Ontology is a key technique that gives the capability of annotating semantics, and provides a common and straightforward infrastructure for formal definition of resources on the domain of concern [40]. Ontologies are described by a 5–tuple

$$(C, H^C, R, H^R, I), \quad (1)$$

Where C represents a set of classes or concepts, and is sorted by a hierarchy defined by $H^C$, and R represents a set of relations between two classes, where $R_i \in R$, and $R_i \rightarrow C \times C$. $H^C$ and $H^R$ depict hierarchies of concepts and relations respectively [9]. Finally $I$ represents individuals of classes (C) or relations (R).

Our desired target of expressivity is SHOIN(D)[1], and we have used OWL–DL as the underlying storage language. Expressivity level of SHIOIQ[2] is slightly more expressive description logic than SHOIN(D), the description logic underlying OWL–DL [32]. Unlike, OWL–Lite, OWL–DL has the capability to reason and infer over knowledge bases. Also unlike OWL–Full, reasoning using description logic over OWL–DL is a decidable problem.

III. NETWORK ATTACK ONTOLOGY

In this section, we describe the methods that have is applied to generate ontological representation from sources like Capec, CWE, and CVE, processes take place to do inference over that ontology, and internal structures which make all these happen from sensory inputs to the prediction results.

In the following sections we describe different pathways to generate final ontology based on two different basic kernel ontologies. At the end, we merge those two domain–level ontologies and use a single knowledge base as an input to the prediction phase. To avoid redundancy, we used unique identifier provided by Capec, CWE, and CVE. In addition, a lexical ontology is provided to translate concepts of these two ontologies to each other.

---

[1] SHOIN(D) means ontology includes attributive language, complex concept negation, transitive property, role hierarchy, nominals , inverse property, cardinality restriction, and uses datatypes.
[2] SHIOIQ is capable of using qualified cardinality restrictions in addition to SHOIN(D) features.

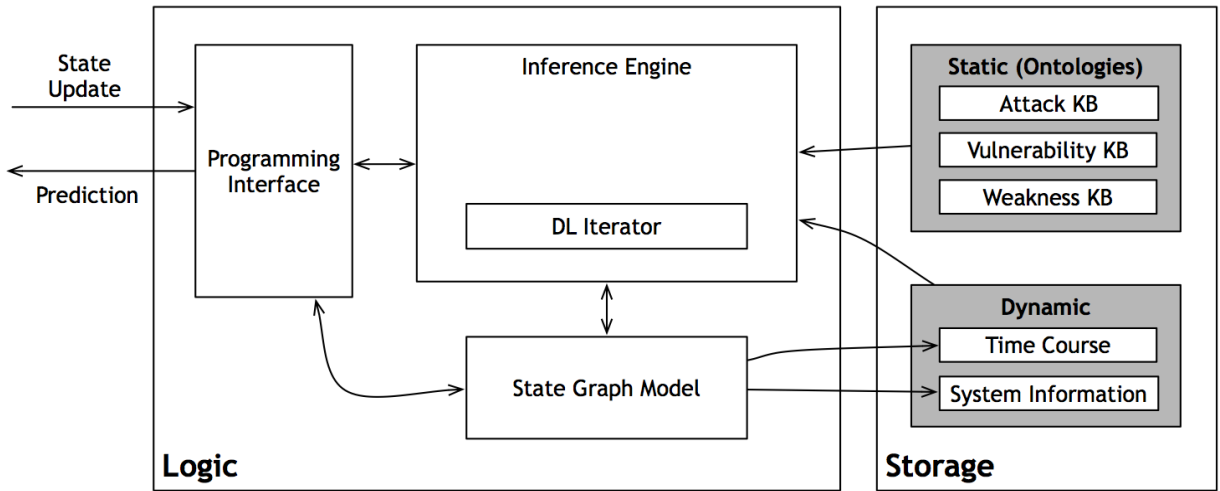

Figure 1. General Architecture of Proposed System for Attack Detection

To provide a basic structure for the growing knowledge base of target system, a translator has been designed, which takes some standard attack taxonomies as inputs, extract semantic structures, and gives multiple ontologies as outputs. The translator is actually a rule–based parser that extracts concepts and structures from taxonomies. It also checks multiple taxonomies for well-matched mutual relations of their concepts (e.g. relations between a single attack pattern from Capec taxonomy and multiple vulnerabilities from CVE taxonomy). The output directed graph is then used as a base point for the knowledge base of attacks where individuals are imported into the knowledge base according to the system events and packet logs. Beside the generated ontologies, higher level domain ontology to complete those events and logs has been designed by hand, so concepts like IP address, ports, and other related system entities will be covered during inference process. In the following sections structures of those ontologies and architecture of the translator will be described in detail.

The first method to generate an ontology and apply prediction methods over it is to model the domain of network attacks by splitting concepts according to their semantics. To split semantically, a basic core ontology consisting of basic attack concepts has been provided. In this method we extend this ontology based on extracted concepts and relations from semi-informal sources like Capec, CWE, CVE, and OSVDB taxonomies.

Three basic classes have been adopted for describing attack domain concepts, 1) Attack Patterns that define the properties of a single attack, 2) Weaknesses which define all weaknesses causing attacks to happen at the first place, and 3) Vulnerabilities.

### A. Complement Domain Ontology

In addition to each attack's properties and relations to vulnerabilities and weaknesses, there are complement concepts which is necessary for knowledge base to model the domain of network attacks properly. These concepts are more related to the system domain and how we model network attacks in general. Fig. 2 shows some parts of the ontology including those complement concepts used for data leakage attacks. Natural language descriptions of each attack in Capec, CVE, and CWE are the sources of such a structure for each attack pattern.

To keep track of attack steps and system properties, final ontology includes a storage unit to store time, data and other properties of a multi–phase attack. Figure x illustrates overall architecture of proposed system for predicting network attacks.

### B. Linking Data Sources and Knowledge Bases

Two methods have been applied to generate ontology from information and data sources. The first method employ rule-based automatic translation to convert data sources like Capec to a mid-level representation, and then automatically generate an ontology by extending a core ontology, and the other method is to create and manage an ontology by extending a core ontology manually.

Outputs of both methods are then checked for its consistency and removing conflicts.

### C. High-level Attack Patterns

Each attack has three essential parts, the prerequisites which includes necessary weaknesses and vulnerabilities, the process which describes all the steps have to be taken to complete the attack and consequence which describes result of the attack.

### D. Weaknesses and Vulnerabilities

Weaknesses and vulnerabilities have been extracted automatically from some standard human reviewed data sources such as CWE and CVE. For each weakness in CWE, a class representing that weakness is created. The hierarchy of weakness ontology ($H^C$) is based on the structure of CWE dataset. Each weakness has a property which points to the related attack. Another ontology is also created by using the same method for vulnerabilities. Both generated ontologies then are merged into the attack ontology. To avoid duplicated entities during replacing common properties with a real class or individual names, unique identifiers were used while generating each ontology from source datasets.

### E. Intermediate Graph Model

Before converting our ontology to a standard language like OWL/RDF, an intermediate model is generated. This model is a bidirectional graph, where each node represents a concept, and each edge represents a relation between two concepts.

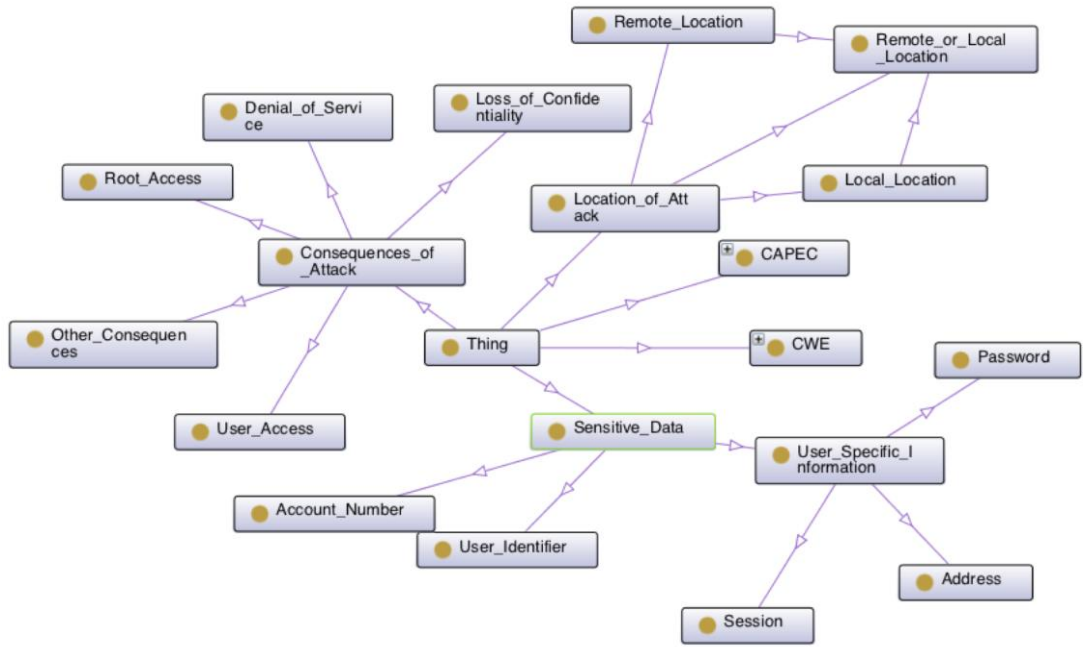

Figure 2. Complement ontology for data leakage attacks

## IV. PREDICTION VIA REASONING

Two different approaches have been employed to detect possible pathways to a network attack. First we use reasoning techniques over an ontology to detect potential attack patterns, and then send the results to a simulation system, to update an internal state, and list possible pathways which can lead to an intrusion. Parser checks if a mapping exists between two entities $E_i$ and $E_j$ respectively, then it adds the equivalent relation to the intermediate graph model.

When the ontologies are imported into the knowledge base, several rules are generated by inference from the chain of implications. From this point the knowledge base is ready to receive instances of concepts of the ontologies. These instances are imported into the knowledge base through OWL/RDF triple syntax. Triple syntax of an instance is in the form of {SUBJECT; PREDICATE; OBJECT}. To query the knowledge base same syntax is used, where subject is a class, predicate is a relation, and object is either a class or a literal. The difference between these two sentences is that in query syntax at most two of three fragments have to be defined as variable by a "?" symbol. If there are some facts in knowledge base, or derived facts resulting from the inference process that match the query pattern, the values of those triples will be provided as the output of reasoning over the knowledge base. Fig. 4 depicts process of feeding data into the knowledge base, and reasoning against collected triples in the knowledge base.

Because of ontological notion of OWL language which is used to represent the knowledge base, different levels of granularity to query the knowledge base are possible. Generated ontologies provide various types of query with varying levels of granularity. The following queries depict three different levels to look up the knowledge based for the existence of a specific attack, all attacks of a specific type, and instances of an attack which has been caused by a specific weakness. Due to the use of SPARQL language for query interface, more complex types of queries are also possible, which is beyond the goals of this paper.

To provide a solid backbone to define rules for attack patterns, four main query types have been implemented: 1) Query for a specific attack, which returns possible individual triples on a specific network node for a specific attack type, including additional individuals required to describe an attack (e.g. attack steps), 2) Query for attacks of a specific attack type, which returns those individual triples defining attack patterns on all nodes in the network which are related to a specific attack (this type of queries helps to identify attacks that are distributed and take place in multiple nodes, e.g. Mitnick attack), 3) Query for all attacks caused by a specific weakness, to provide the reasoner with possible attacks caused by a discovered weakness class or individual (as a result, administrators can decide about the priority of resolving a weakness based on the possibility of potential attacks), and 4) Query for attacks caused by a specific vulnerability class or individual.

The prediction phase employs reasoning techniques and iterate over all returned triples to detect potential attacks. In this phase a query based on observed events, weaknesses, vulnerabilities, and attacks is generated by the system in SPARQL format. Then results from knowledge base are gathered and grouped based on their similarities. The final step of prediction is matching those triples with current system state, and returning those triples which are mostly appropriate to happen based on the current network state. The aim of this multi–phase technique is to firstly overcome shortage of the ontology to keep track of dynamic parameters of events (e.g time) which are important in multi–step attacks, and secondly simplify queries.

The main idea behind prediction is to utilize a set of Semantic Rule Language (SWRL) rules that represent vulnerabilities, weakness and other prerequisites of each attack in a single place.

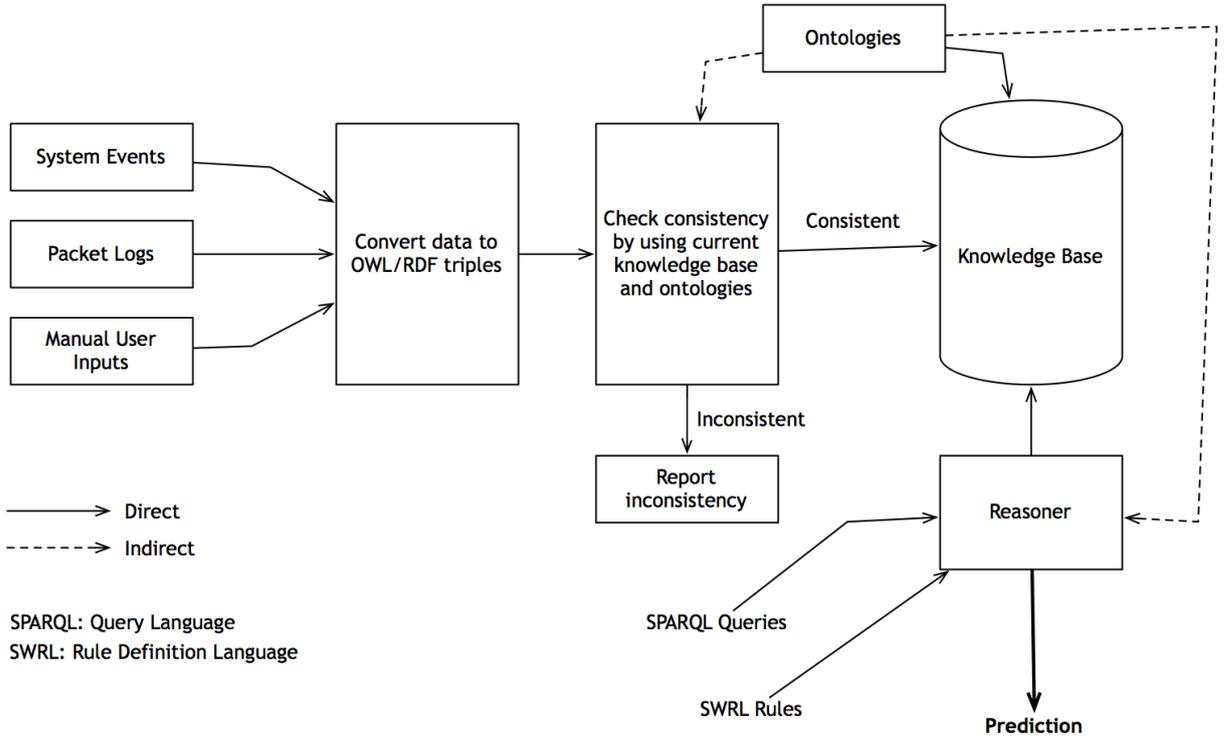

Figure 3. Architecture of feeding observed data into the knowledge base

Formula 2 illustrates how a network node (?s) is marked as **Vulnerable**, and formula 3 represents overall pattern of attack detection formula. Predictability can be achieved by changing ∧s (logical ANDs) of vulnerabilities to ∨s (logical ORs) of them in formula 3. Formula 4 depicts such prediction rule.

## V. EVALUATION OF PRODUCED ONTOLOGY

To evaluate and test our ontologies, we applied three different approaches. We gave our ontologies to three different human experts in network attack domain to evaluate the quality of the ontologies. Then various quantity parameters have been extracted for quantitative evaluation. And finally we asserted them into the knowledge base and ran queries against the knowledge base. In the second phase of evaluation some predefined events have been asserted into the knowledge base, which are prerequisites of multiple attacks, so test prediction unit could be tested by predicting those desired attacks.

Formulas 2,3, and 4. Main Prediction Rules Based on Vulnerabilities

$$\text{System}(?s) \wedge \text{hasWeakness}(?s, ?w) \wedge \text{relatedTo}(?w, ?v) \rightarrow \text{Vulnerable}(?n) \quad (2)$$

$$\text{Vulnerable}(?n) \wedge \text{hasVulnerability}(?n, V_1) \wedge \ldots \wedge \text{hasVulnerability}(?n, V_n) \\ \rightarrow \text{UnderAttackSystem}(?n) \quad (3)$$

$$\text{Vulnerable}(?n) \wedge (\text{hasVulnerability}(?n, V_1) \vee \ldots \vee \text{hasVulnerability}(?n, V_n)) \\ \rightarrow \text{UnderPotentialAttackSystem}(?n) \quad (4)$$

### A. Quantitative Evaluation

To evaluate and describe certain aspects of designed ontology, several metrics have to be extracted from ontology and knowledge based structures. These metrics provide comparability between two versions of the same ontology during the ontology's lifecycle, make it possible to compare and evaluate changes in the domain described by the ontology. Unlike qualitative parameters, the metrics described in this section cannot be used to evaluate if an ontology is good or bad, but they can be used to track changes in quantitative characteristics such as richness over the ontology's lifecycle. Since ontologies are usually designed with their planned applications in mind, they have different characteristics from each other. As a result of these different approaches of ontology design, those ontologies that are aimed for the same task are comparable with each other.

Quantitative metrics are divided into two groups: those related to ontology, and those related to knowledge base (ontology and individuals). The first group evaluates design and potential of the ontology in representing desired knowledge, and the second group evaluates effective utilization of the ontology to model individuals of the domain. Following a description of both groups is presented. These quantitative parameters can be used to improve certain aspects of designed knowledge representation of the domain. As an example of this kind of improvement, one can improve the ontology's capability to model each single attack's properties, or track attack categorization changes during knowledge based lifecycle.

**Ontology Metrics**: These metrics indicate richness, width, depth, and hierarchical level of an ontology design [27]. The most important metrics of this group are:

1) Object Properties Richness, which reflects the diversity of non–inheritance relationships, and is represented as the number of objects and data properties between classes compared to all possible relations (including hierarchical relationships like is-a).

2) Inheritance Richness, is defined as the average number of subclasses per class.

3) Data Properties Richness, is defined as the average number of functional object properties in addition to data properties per class.

**Knowledge Base Metrics**: Beside these design related metrics of ontologies, it's important to measure how the ontology is utilized to model real world individuals such as attacks. Knowledge base metrics describe the quality of an ontology to assign class or properties to the individuals and cover concepts of the domain. Below is the list of knowledge based metrics:

4) Class Richness, describes how instances are distributed across classes. This is defined as the number of non–empty classes (those with at least one individual), divided by the total number of classes defined in the ontology.

5) Class Connectivity, gives an indication of what classes are important in the ontology, based on the relationships between individuals of all classes (known as individual graph or knowledge base graph). This metric is defined as the number of object properties between individuals of a class, and individuals belong to other classes.

6) Class Importance, is calculated by number of individuals belong to classes at the inheritance sub tree rooted at the chosen class, divided by the total number of individuals.

7) Individual Graph Components, shows how individuals are related to each other, and defined as the number of connected components of the knowledge base graph.

8) Object Properties Richness, reflects how much of object properties for a class in the ontology are actually used at the individuals level. This metric is defined as the number of object properties that are being used by individual $I_i$ of class $C_i$, compared to the number of all object properties that are defined for $C_i$ at the ontology level.

[28] and [15] list even more metrics that evaluate various aspects of ontologies, and knowledge bases, which are beyond scope of this paper.

In addition to these metrics, the generated ontologies have been evaluated using the following metrics:

- Quantity of class nominated,
- medium of properties ($P_o$),
- is-a object property level of the ontology,
- class bigger rank of the is-a object property, and
- class bigger rank of all-part object property.

Table 1 lists various quantitative metric parameters of produced attack ontology.

## VI. EVALUATE INFERENCE (USE CASE)

In this section a use case for evaluating prediction rules (aside from graph structure of ontology) is explored. To evaluate reasoning capability of produced ontology (including attack prediction rules) and knowledge base, we initiated some attacks on network nodes by simulating logs and events for each. Triples representing simulated logs and events then asserted into the knowledge base. Figure 4 shows the rule is used to mark a system as vulnerable, for a data leakage attack so–called JavaScript Hijacking or JSON Hijacking attack with Capec ID of 111. Capec describes this attack as "*An attacker targets a system that uses JavaScript Object Notation (JSON) as a transport mechanism between the client and the server (common in Web 2.0 systems using AJAX) to steal possibly confidential information transmitted from the server back to the client inside the JSON object by taking advantage of the loophole in the browser's Single Origin Policy that does not prohibit JavaScript from one website to be included and executed in the context of another website*" [2]. These rules and those simulated logs and events asserted into the knowledge base. Then by querying subclasses of UnderPotentialAttackSystem, it could be seen simulated system node is among them (as an under attack node).

As it can be seen, even simple attacks need multiple independent long rules. Unlike traditional methods like taxonomies or attack languages, the overall technique of using ontology and SWRL (which tends to store rules as concepts and properties in OWL format rather than different format) makes them maintainable and easy to update for new attacks.

As a use case scenario for testing multi–phase attacks, Mitnick attack set of rules asserted into the knowledge base and system nodes tested on it. As a result of feeding logs and events of simulated Mitnick attack, in all steps of this attack the prediction system predicted it successfully by marking respective nodes as Vulnerable and UnderPotentialAttackSystem.

| Parameter | Value |
| --- | --- |
| Object Properties Richness (Ontology) | %6.7 |
| Inheritance Richness (Ontology) | 5.4 |
| Data Properties Richness (Ontology) | 8.2 |
| Number of Classes (Attacks) | 394 |
| Medium of $P_o$ | 28/394 |
| is-a Relations Level | 7 |
| Class Bigger Rank of is-a | $AttackPattern(7)$, $SystemObject(5)$ |
| Class Bigger Rank of part-of | $AttackPattern(3+9=12)$ |

Table 1. Quantitative Evaluation Metrics For Attack Domain Ontolog

$$System(?s) \land hasWeakness(?s, CWE345) \land hasWeakness(?s, CWE346) \\ \land hasWeakness(?s, CWE352) \rightarrow Vulnerable(?s) \quad (5)$$

Figure 4. JavaScript hijacking attack (CAPEC–111) prediction rules

## VII. CONCLUSION

These results demonstrate that the ontology-driven prediction methods are a promising approach for the network attack domain. Since network attacks are mostly described in detail by various semi-informal datasets, the following future researches are possible in the field of network attacks: ontology-driven intercommunication between heterogeneous intrusion detection agents with different languages and models, knowledge-based intelligent intrusion detection, and interdisciplinary methods to predict and prevent network attacks by merging multiple knowledge bases from different domains as well as detect and resolve weaknesses and vulnerabilities of a network node by reasoning over past attacks' information.

Due to the generalized conceptualization of produced ontology, its applications can be extended to other attack-related problems including, but not limited to attack report unification, correlation, and recognition. For these kinds of applications more detailed attacks, vulnerabilities, and weaknesses can be imported into the further ontology. To improve richness of the ontology, informal knowledge in natural language sources could be extracted from various informal sources using natural language processing techniques. Prediction technique presented in this paper can be extended to consider more parameters (e.g. various network behaviors and anomalies). In addition, since forward-chaining reasoning is much faster for small set of axioms, proposed logic-base technique can be extended to employ machine learning classification techniques to reduce axioms used as base for inference.

## ACKNOWLEDGMENT

This research has been supported by Research Institute for ICT (ex ITRC).

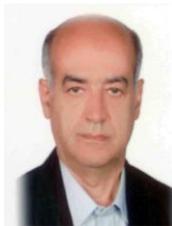

**Ahmad Salahi** was born in Tehran, Iran, on Feb.10.1947. He received his B.Sc. degree in electrical engineering from Tehran University, Iran, his M.Sc. degree from Kansas University U.S.A in 1974 and his Ph.D. degree from Purdue University West Lafayette Indiana, U.S.A in 1979, all in electrical engineering. At present, he is a senior project manager in Network Security Department in Iranian Research Institute for ICT (ex. ITRC). His research interests are network security, switching and routing.
.

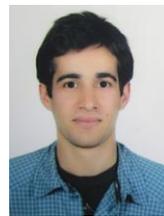

**Morteza Ansarinia** was born In Sari, Iran on 1985. He started his B.Sc. degree in Computer Engineering at Amirkabir University in 2003, his M.Sc. degree in Artificial Intelligence at Shahid Beheshti University in 2008, and his Ph.D. degree in Compuational Intelligence at Uninova Institute/CA3 Lab in 2010. In 2010, he attended European Space Agency-Funded project for knowledge discovery in very large datasets. He joined Iranian Research Institute for ICT (ex. ITRC) as resident researcher of knowledge management's methods starting from February 2012. His research intrests include semantic information processing, knowledge discovery, and distributed intelligence.